\documentclass{article}

\usepackage{PRIMEarxiv}

\usepackage[utf8]{inputenc} 
\usepackage[T1]{fontenc}    
\usepackage{hyperref}       
\usepackage{url}            
\usepackage{booktabs}       
\usepackage{amsfonts}       
\usepackage{nicefrac}       
\usepackage{microtype}      
\usepackage{lipsum}
\usepackage{fancyhdr}       
\usepackage{graphicx}       
\graphicspath{{media/}}     

\title{Integrating Edge-AI in Structural Health Monitoring domain
\thanks{\textit{\underline{Citation}}: 
\textbf{Authors. Title. Pages.... DOI:000000/11111.}} 

}

\author{
  Anoop Mishra \\
  University of Nebraska \\
  6001 Dodge Street\\
  Omaha, Nebraska \\
  USA - 68182\\
  \texttt{amishra@unomaha.edu} \\
   \And
   Gopinath Gangisetti \\
   University of Nebraska \\
   6001 Dodge Street\\
   Omaha, Nebraska \\
   USA - 68182\\
   \texttt{ggangisetti@unomaha.edu} \\
   \And
   Deepak Khazanchi \\
   University of Nebraska \\
   6001 Dodge Street\\
   Omaha, Nebraska \\
   USA - 68182\\
   \texttt{khazanchi@unomaha.edu} \\
}

\begin{document}
\maketitle

\begin{abstract}
\textbf{Structural health monitoring (SHM)} tasks like damage detection are crucial for decision-making regarding maintenance and deterioration. For example, \textbf{crack detection} in SHM is crucial for bridge maintenance as crack progression can lead to structural instability. However, most AI/ML models in the literature have low latency and late inference time issues while performing in real-time environments. This study aims to explore the integration of edge-AI in SHM domain for \textbf{real-time bridge inspections}. Based on edge-AI literature, its capabilities will be valuable integration for a real-time decision support system in SHM tasks such that real-time inferences can be performed on physical sites. This study will utilize commercial edge-AI platforms, such as Google Coral Dev Board or Kneron KL520, to develop and analyze the effectiveness of edge-AI devices. Thus, this study proposes an \textbf{edge AI framework for the structural health monitoring domain}. An edge-AI-compatible deep learning model is developed to validate the framework to perform \textbf{real-time crack classification}. The effectiveness of this model will be evaluated based on its accuracy, the confusion matrix generated, and the inference time observed in a real-time setting. 
\end{abstract}


\section{Introduction}
The increasing traffic loads and threat of structural fatigue make it crucial to monitor the health condition of structural bridges. \textbf{Structural Health Monitoring (SHM)} can play a key role in increasing the lifespan and performance of bridges \cite{gharehbaghi2022novel}. The optimal system for monitoring, managing, and assessing the condition of bridges is required for performing tasks like bridge \textbf{maintenance and deterioration}. Periodic and continuous monitoring via bridge deck inspections is necessary for structural health monitoring (SHM) \cite{liu2019computer,yang2015thin}. Advances in technology like wireless sensor networks and artificial intelligence have allowed bridge structural health monitoring more effective. In the past, AI approaches including machine learning, computer vision, and robotics along with data mining and image processing techniques are utilized to address the SHM challenges \cite{reghukumar2021crack,flah2021machine,guo2020data,gandhi2018hidden}. Especially deep learning approaches are broadly utilized in the past for detecting and segmenting bridge components \cite{cui2021intelligent, tan2019deep,yang2020crack}. Results obtained through these computational approaches have shown immense potential to support health monitoring tasks such as crack detection, segmentation of bridge components, UAV inspection of bridges, and damage detection for bridge structure \cite{zhang2021structural,reghukumar2021crack,guo2020data,gandhi2018hidden}. However, real-time inferences in SHM, especially \textbf{crack detection} is challenging based on several factors like a wide range of various complex backgrounds and crack-like features. \par

Edge-AI allows a process for artificial intelligence inference and computation on-device rather than on cloud servers or network connections. Edge-AI optimizes the response time and accelerates AI computation when data is generated and processed for inference on-device \cite{zhou2019edge}. Zhou et al 2019 \cite{zhou2019edge} suggests that this ability of edge-AI platforms has advantages as follows-
\begin{enumerate}
    \item Increase security and privacy,
    \item Reduce need for data storage and bandwidth,
    \item Reduce latency,
    \item Increase stability
    \item Improves cost optimization for AI integration for dedicated tasks, and 
    \item Improves real-time inference time
\end{enumerate}

\begin{figure*}
\begin{center}
\includegraphics[width=1.0\textwidth, height = 10cm]{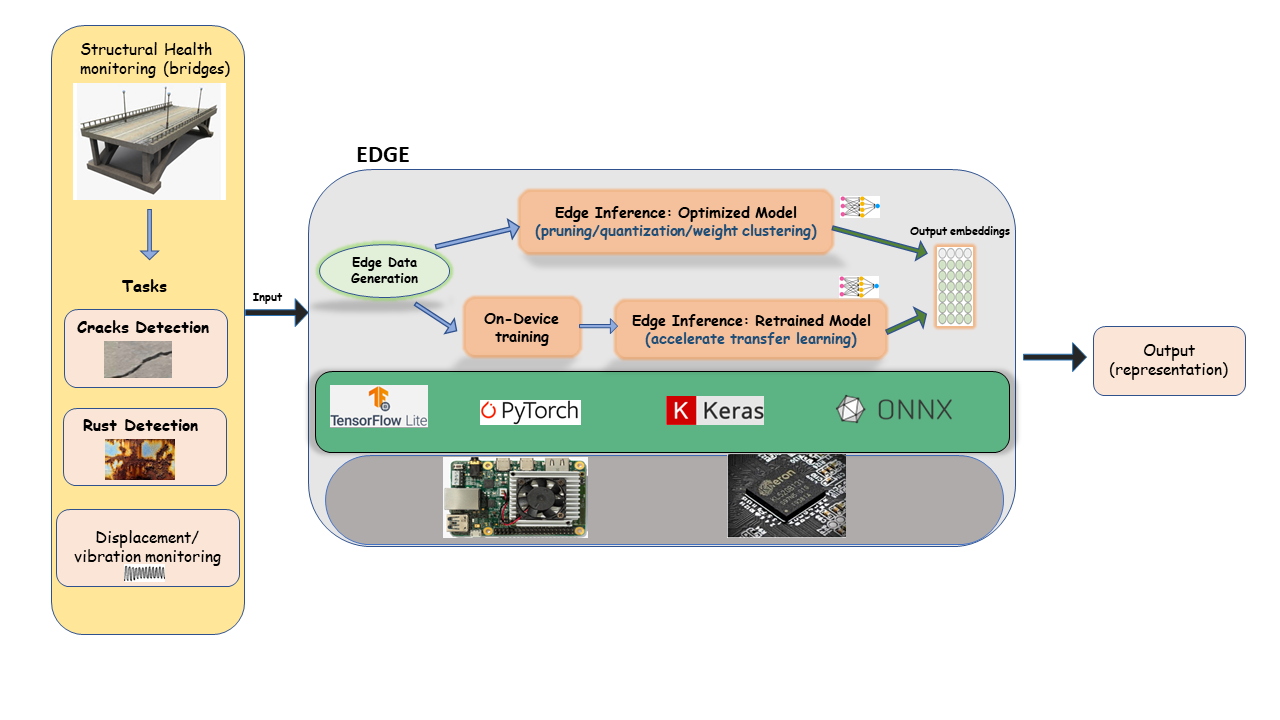}
\end{center}
\caption{Proposed edge-AI framework in the structural health monitoring domain}
\label{framework}
\end{figure*}

\subsection{Research Goals and Proposed approach}
 In this research study, drawing from artificial intelligence (AI), neural network process acceleration, and edge computing literature, the study aims to investigate edge-AI integration in structural health monitoring tasks. The objective is to develop a lightweight neural network model and to optimize neural network response time performance in real time. To accomplish this, we are utilizing \textbf{Kneron KL520} platform which includes a neural processing unit (NPU), an AI system-on-chip(SoC) hardware. \par 

To address the inherent challenges for real-time inference an edge-AI framework for SHM domain is proposed. This paper introduces a novel framework to generate lightweight optimized models on the edge within the SHM domain. Figure \ref{framework} describes the edge-AI-SHM framework. The framework design explains that the neural network for edge-AI chips can be developed using an optimized model that includes pruning, quantization, and weight clustering approach or by on-device training where the transfer learning approach is used to fine-tune the existing edge neural network model. Crack detection as an SHM task is experimented on the Kneron KL520 platform to validate the proposed framework. More details of the experiments can be found in section \ref{sec:methods}. In summary, the technical contributions are as follows:
\begin{enumerate}
    \item Proposed a novel edge-AI framework for SHM domain.
    \item Developed and deployed the optimized CNN model for classifying cracks on image data in a real-time setting
    \item Optimized the CNN inference time for identifying cracks from 3 seconds (mean) to 20 milliseconds (mean)
\end{enumerate}

This study will help researchers to practice all-on-device edge AI as a technology for real-time applications and research purposes in the fields such as robotics, smart surveillance, and automation in the domain of structural health monitoring.

\section{Methods} \label{sec:methods}
\subsection{Dataset}
\begin{figure}[ht]
\centering
\includegraphics[width=0.23\textwidth, height = 3cm]{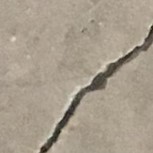}
\includegraphics[width=0.23\textwidth, height = 3cm]{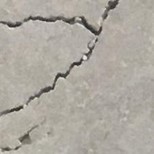}
\includegraphics[width=0.23\textwidth, height = 3cm]{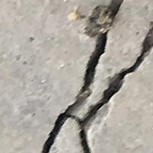}
\includegraphics[width=0.23\textwidth, height = 3cm]{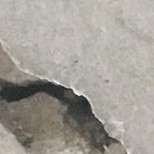}
\caption{Sample images from positive class of the dataset from Caglar et al. \cite{ccauglar2019concrete}}
\label{positive_cracks}
\end{figure}

The public crack dataset from Caglar et al. 2019 includes concrete building crack images from METU campus buildings \cite{ccauglar2019concrete}. The dataset consists of two class labels, i.e., “Positive” and “Negative”. Each class label contains 20000 images with a total of 40,000 images of cracks and non-cracks. The image characteristics of the images describe that images contain 3 color channels (RGB) with 227x227 dimensions. Figure \ref{positive_cracks} shows the sample positive cracks from the dataset.

\subsection{Kneron KL520} \label{KL}
This study is using Kneron KL520 chip series to develop edge-AI based optimized model. More details of Kneron KL520 can be found on this URL: \url{https://www.kneron.com/page/soc/}. Based on \cite{kne2023edge}, brief specifications of the chip are as follows-
\begin{enumerate}
    \item Single-on-chip (SoC) design platform
    \item 32MB SDRAM (16-bit LPDDR2-1066)
    \item Neural Processor Unit(NPU) for accelerating neural network process
    \item CPU: Two ARM Cortex-M4, one for system control and the other as a AI co-process for NPU
    \item Maximum Frequency: 300 MHz
\end{enumerate}

Although the device supports popular CNN models like VGG16, VGG19, MobileNet, etc., but it has support for a limited set of operators to design a CNN architecture. This restricts designing the model architecture specific to this very device. All the details related to the operators can be found on the website (\url{https://www.kneron.com/page/soc/}).

\subsection{Neural network model architecture}
This study aims to develop a cracks image classification model based on \textbf{Convolutional neural network (CNN)} architecture. The model architecture consists of an input layer of 224x224x3 dimension, 6 convolutions with max-pooling layers, 2 dense layers, and an output layer with a Softmax activating function. The development stack for this research study includes Python3 with AI frameworks like Tensorflow lite, Keras, and ONNX for embedded model conversion. 

\subsection{Experiments: Overview}
\subsubsection{Training:}
\begin{figure*}
\begin{center}
\includegraphics[width=1.0\textwidth, height = 10cm]{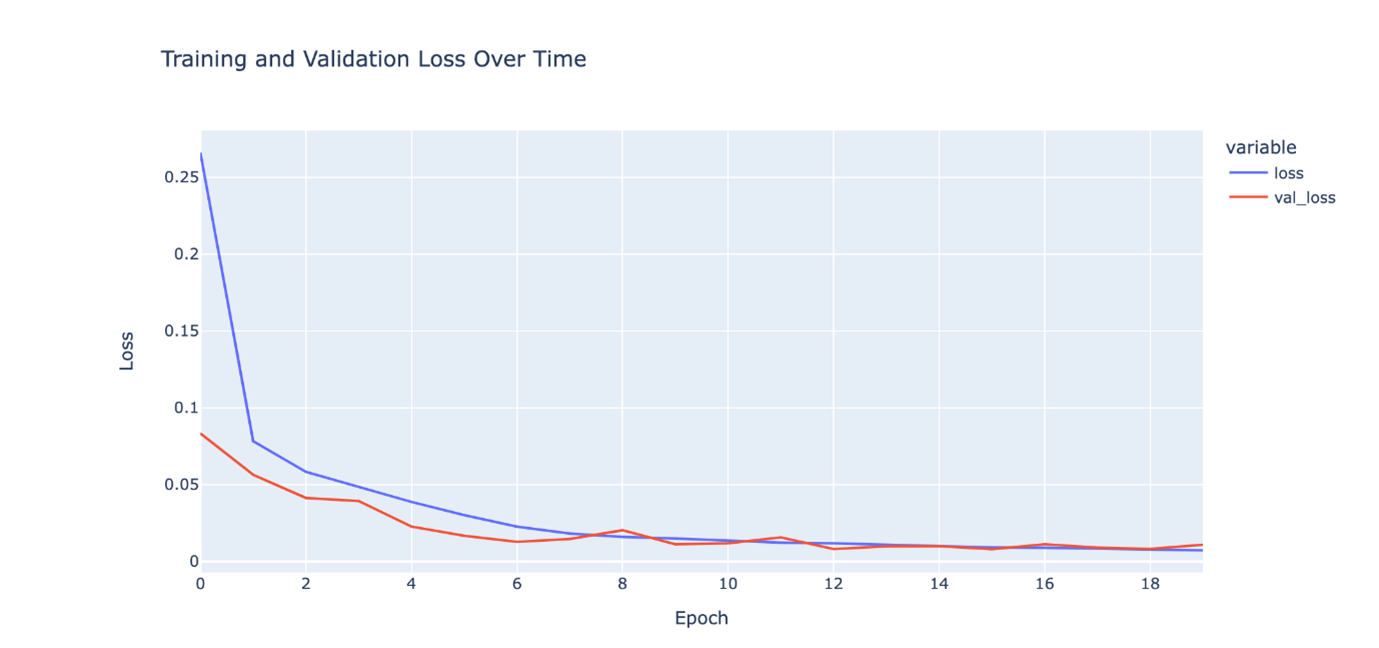}
\end{center}
\caption{Training performance of the main model}
\label{performance}
\end{figure*}

The model training is designed based on training and validation sets, with a split of 80-20 based on 40000 images. The training is iterated for 20 epochs. The results of the training and validation accuracy are both very high, at 99.54\% and 99.52\% respectively. This means that the model is very accurate in classifying the images of the training and validation data. Testing the inference for the same 40000 images on simulation of the Kneron KL520 chip resulted in 92.4\% accuracy. The testing inference is discussed in section \ref{test_instance}. Figure \ref{performance} shows the training performance of the developed model.

\subsubsection{Optimized model generation for Chip compability}

To generate an optimized model and test it on the Kneron chip, there are two major steps: 
\begin{enumerate}
    \item Make sure the operators used in the original model are supported by the device and convert the original model to ONNX file format for further use.
	\item Testing the model performance using simulation and getting a view of the results before deploying the model to the device.
\end{enumerate}

 As mentioned in section \ref{KL}, the edge device supports limited operators. The softmax activation function is one of the operators not supported by the Kneron NPU. Thus the softmax layer had to be removed before saving the model for device format conversion. \par

\textbf{ONNX conversion:} ONNX is a file format for storing deep learning models. It is useful for edge computing because it allows models to be deployed to devices that may not have access to a full deep learning framework. ONNX also supports a number of different hardware architectures, so models can be deployed to a variety of devices. \par

\textbf{Quantization:} It is a process of transforming the number of bits that represent a number. In deep learning, quantization is often used to reduce the size and complexity of models, making them faster and more efficient. In this study, the general model was transformed into an 8-bit model.

Before generating an optimized model that is supported by the device, the following steps are required:

\begin{enumerate}
    \item Create an optimized onnx file by converting the model from several platforms. Examine the onnx model for verifying operator compatibility. The onnx model is then put to the test, and the output is compared to the original.
	\item Create a fixed-point file using the post-quantization approach on the main (floating-point) model. 
	\item Create a binary file in the \textbf{nef} format (npu executable format) format.
\end{enumerate}

The flowchart in figure \ref{chart} describes the workflow for optimized model development.

\subsubsection{Kneron KL520 inference on test sample} \label{test_instance}
For testing in a real-time setting, the host device (mainframe system with camera capability) captures the image and sends the image to know for inference. Before the image is sent for inference, the image is pre-processed on the host device aligning with the input layer. Also, since the output layer has a softmax activation function removed before NEF format conversion, the post-processing of the raw output generated from KL520 is also done on the host device. The testing sample includes a total of 355 images out of which 255 are of cracks class and 100 are of non-cracks. This test sample includes images from the main dataset and also includes 100 images (both positive and negative) collected manually for real-time testing. The testing accuracy performance is impacted by the sample collected manually, as they belong to different data distributions. They differ in cracks characteristics like length and width, and also in background texture. 

\begin{figure}
\setlength{\fboxsep}{0pt}%
\setlength{\fboxrule}{0pt}%
\begin{center}
\includegraphics[width=0.4\textwidth,, height = 6cm]{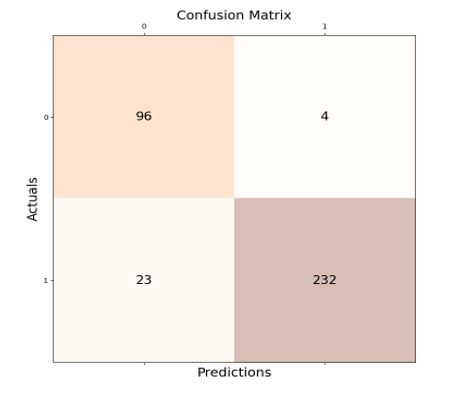}
\end{center}
\caption{Confusion matrix for testing sample}
\label{cf}
\end{figure}

\textbf{Testing performance: } The overall accuracy of the testing results is 92.4\%, which can be counted as imposing. The average inference time per image on the device was 20 milliseconds including pre-processing and post-processing on the host device. The Kneron KL520 chip provides an efficient AI computing performance of 0.35 TOPS per watt. The confusion matrix can be seen in figure \ref{cf}. The testing accuracy can be improved by utilizing on-device training by retraining the model using transfer learning approach as proposed in the edge-AI framework in figure \ref{framework}.

\section{Conclusion}
This study proposed a novel edge-AI framework for the tasks in the structural health monitoring(SHM) domain to perform real-time inference on physical sites. Following the proposed framework, a post-quantization optimized model is developed having CNN architecture. The testing accuracy for the crack image classification is 92.4\%. The inference time is optimized from an average of 3 seconds to 20 milliseconds per image.

\section{Acknowledgement}
This research project belongs to the College of Information, Science and Technology, University of Nebraska at Omaha.

\bibliographystyle{unsrt}  
\bibliography{references} 

\appendix
\begin{figure*}
\begin{center}
\includegraphics[width=0.7\textwidth, height = 20cm]{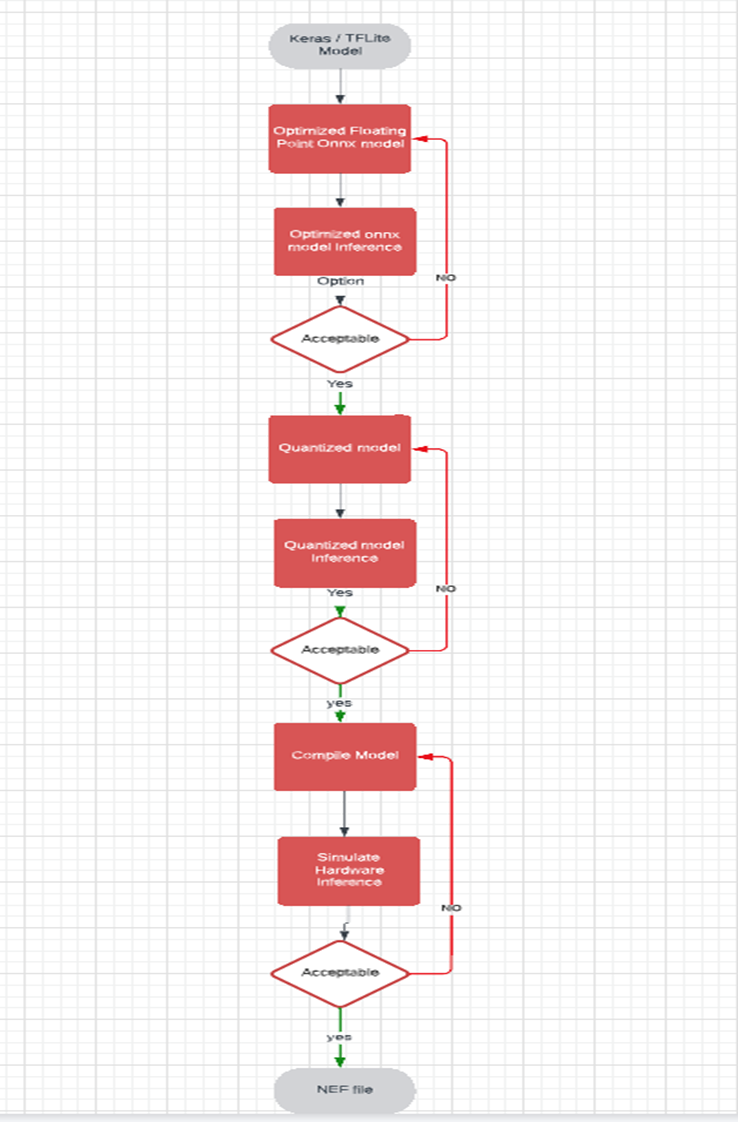}
\end{center}
\caption{Flowchart for developing an optimized model to inference on Kneron KL520}
\label{chart}
\end{figure*}

\end{document}